\documentclass[runningheads]{llncs}
\usepackage[dvipsnames]{xcolor}

\usepackage{graphicx}

 \usepackage{array}
 \usepackage{multirow}
\usepackage{tikz}
\usepackage{comment}
\usepackage{amsmath,amssymb} 
\usepackage{cancel}
\usepackage{booktabs}
\usepackage{caption}
\usepackage{breakcites}
\usepackage[nocompress]{cite}
\usepackage[colorlinks,anchorcolor=Maroon,urlcolor=Maroon,citecolor=Maroon,linkcolor=Maroon,breaklinks=true,pagebackref=true]{hyperref}       
\usepackage[boxed]{algorithm}
\usepackage{algorithmicx}
\usepackage{algpseudocode}
\usepackage{wrapfig}

\let\llncssubparagraph\subparagraph
\let\subparagraph\paragraph
\usepackage{titlesec}
\let\subparagraph\llncssubparagraph

\usepackage[accsupp]{axessibility}  

\usepackage{stmaryrd}
\usepackage{trimclip}

\makeatletter
\DeclareRobustCommand{\shortto}{%
  \mathrel{\mathpalette\short@to\relax}%
}

\newcommand{\short@to}[2]{%
  \mkern2mu
  \clipbox{{.5\width} 0 0 0}{$\m@th#1\vphantom{+}{\shortrightarrow}$}%
  }
\makeatother

\DeclareMathOperator*{\argmax}{arg\,max}

\newcommand{\T}{^\top}

\newcommand{\parhead}[1]{\smallskip \noindent \textbf{#1}}

\newcommand{\eref}[1]{(\ref{#1})}
\newcommand{\secref}[1]{Sec.~\ref{#1}}
\newcommand{\figref}[1]{Fig.~\ref{#1}}
\newcommand{\tabref}[1]{Tab.~\ref{#1}}
\newcommand{\algoref}[1]{Alg.~\ref{#1}}

\titlespacing\section{0pt}{12pt plus 4pt minus 2pt}{4pt plus 2pt minus 2pt}
\titlespacing\subsection{0pt}{10pt plus 4pt minus 2pt}{2pt plus 2pt minus 2pt}
\titlespacing\subsubsection{0pt}{8pt plus 4pt minus 2pt}{0pt plus 2pt minus 2pt}

\begin{document}
\pagestyle{headings}
\mainmatter
\def\ECCVSubNumber{602}  

\title{RelPose: Predicting Probabilistic Relative Rotation for Single Objects in the Wild} 

\titlerunning{RelPose: Probabilistic Relative Rotation Estimation}
%
\author{
Jason Y. Zhang
\and Deva Ramanan
\and Shubham Tulsiani
}

\authorrunning{J.~Y.~Zhang et al.}
%
\institute{Carnegie Mellon University, Pittsburgh PA 15213, USA \\
\email{jasonyzhang@cmu.edu}}
\maketitle

\begin{figure}[b]
    \centering
    \includegraphics[width=\textwidth]{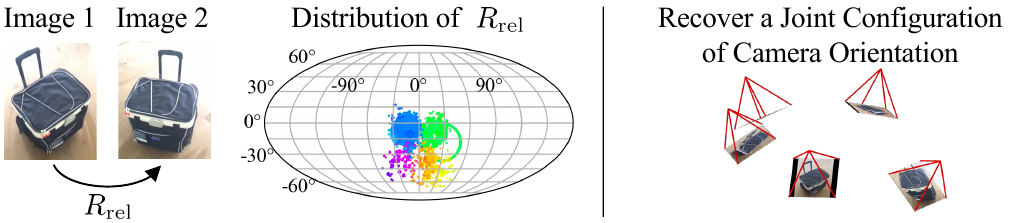}
    \caption{\textbf{Probabilistic Camera Rotation Estimation for Generic Objects}. \textit{Left}: Given two images of the same object, we predict a conditional distribution of relative camera viewpoint (rotation) that effectively handles symmetries and pose ambiguities. \textit{Right}: Given a set of images, our approach outputs a configuration of camera rotations.}
    \label{fig:teaser}
\end{figure}

\begin{abstract}
We describe a data-driven method for inferring the camera viewpoints given multiple images of an arbitrary object.
This task is a core component of classic geometric pipelines such as SfM and SLAM, and also serves as a vital pre-processing requirement for contemporary neural approaches (e.g. NeRF) to object reconstruction and view synthesis. 
In contrast to existing correspondence-driven methods that do not perform well given sparse views, we propose a top-down prediction based approach for estimating camera viewpoints.
Our key technical insight is the use of an energy-based formulation for representing distributions over relative camera rotations, thus allowing us to explicitly represent multiple camera modes arising from object symmetries or views. Leveraging these relative predictions, we jointly estimate a consistent set of camera rotations from multiple images. We show that our approach outperforms state-of-the-art SfM and SLAM methods given sparse images on both seen and unseen categories. Further, our probabilistic approach significantly outperforms directly regressing relative poses, suggesting that modeling multimodality is important for coherent joint reconstruction. We demonstrate that our system can be a stepping stone toward in-the-wild reconstruction from multi-view datasets. The project page with code and videos can be found at \href{https://jasonyzhang.com/relpose}{jasonyzhang.com/relpose}.

\end{abstract}

\section{Introduction}

Recovering 3D from 2D images of an object has been a central task in vision across decades. Given multiple views, structure-from-motion (SfM) based methods can infer a 3D representation of the underlying instance while also associating each image with a camera viewpoint. However, these correspondence-driven methods cannot robustly handle sparsely sampled images that minimally overlap, and typically require many ($>$20) images for a 360-degree 3D inference. Unfortunately, this requirement of densely sampled views can be prohibitive---online marketplaces often have only a few images per instance, and a user casually reconstructing a novel object would also find capturing such views tedious.  Although the recently emerging neural 3D reconstruction techniques also typically leverage similarly dense views, some works have shown promise that a far smaller number of images can suffice for high-quality 3D reconstruction. These successes have however still relied on precisely~\cite{pix3d} or approximately~\cite{lin2021barf, rebain2022lolnerf,zhang2021ners,goel2021differentiable} known camera viewpoints for inference. To apply these methods at scale, we must therefore answer a fundamental question—\emph{given sparsely sampled images of a generic object, how can we obtain the associated camera viewpoints}?

Existing methods do not provide a conclusive answer to this question. On the one hand, bottom-up correspondence-based techniques are not robustly applicable for sparse-view inference. On the other, recent neural multi-view methods can optimize already known approximate camera poses but provide no mechanism to obtain these to begin with. In this work, our goal is to fill this void and develop a method that, given a small number of unposed images of a generic object, can associate them with (approximate) camera viewpoints. Towards this goal, we focus on inferring the camera rotation matrices corresponding to each input image and propose a top-down approach to predict these. However, we note that the `absolute' rotation is not well-defined given an image of a generic object—it assumes a `canonical' pose which is not always known a-priori (e.g. what is an identity rotation for a pen? or a plant?). In contrast, the {\em relative} rotation between two views is well-defined even if a canonical pose for the instance is not. Thus, instead of adopting the common paradigm of single-image based pose prediction, we learn  to estimate the relative pose given a pair of input images. We propose a system that leverages such pairwise predictions to then infer a consistent set of global rotations given multiple images of a generic object.

A key technical question that we consider is regarding the formulation of such pairwise pose estimation. Given two informative views of a rotationally asymmetric object, a regression-based approach may be able to accurately predict their relative transformation. The general case however, can be more challenging---given two views of a cup but with the handle only visible in one, the relative pose is ambiguous given just the two images.  To allow capturing this uncertainty, we formulate an energy-based relative pose prediction network that, given two images {\em and} a candidate relative rotation, outputs an energy corresponding to the (unnormalized) log-probability of the hypothesis.  This probabilistic estimation of relative pose  not only makes the learning more stable, but more importantly, provides a  mechanism to estimate a \emph{joint distribution} over viewpoints given multiple images. We show that optimizing rotations to improve this joint likelihood yields coherent poses given multiple images and leads to significant improvements over naive approaches that do not consider the joint likelihoods.

We train our system using instances from over 40 commonplace object categories, and find that not only can it infer accurate (relative) poses for novel instances of these classes, it even generalizes to instances from unseen categories. Our approach can thus be viewed as a stepping stone toward sparse-view 3D reconstruction of generic objects; just as classical techniques provide precise camera poses that (neural) multi-view reconstruction methods can leverage, our work provides a similar, albeit coarser, output that can be used to initialize inference in current (and future) sparse-view reconstruction methods. While our system only outputs camera rotations, we note that a reasonable corresponding translation can be easily initialized assuming object-facing viewpoints, and we show that this suffices in practice for bootstrapping sparse-view reconstruction.

\begin{figure}[t]
    \centering
    \includegraphics[width=\textwidth]{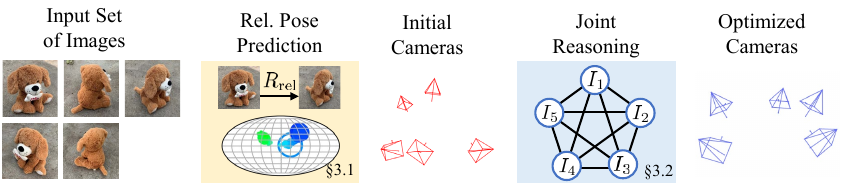}
    \caption{\textbf{Overview.} From a set of images, we aim to recover corresponding camera poses (rotations). To do this, we train a pairwise pose predictor that takes in two images and a candidate relative rotation and predicts energy. By repeatedly querying this network, we recover a probability distribution over conditional relative rotations (see \secref{sec:method_pair}). We use these pairwise distributions to induce a joint likelihood over the camera transformations across multiple images, and iteratively improve an initial estimate by maximizing this likelihood (see \secref{sec:method_joint}).}
    \label{fig:overview}
\end{figure}

\section{Related Work}

\parhead{Structure-from-Motion (SfM).}
At a high level, structure-from-motion aims to recover 3D geometry and camera parameters from image sets. This is done classically by computing local image features~\cite{harris1988combined,lowe2004distinctive,bay2006surf,tola2009daisy}, finding matches across images~\cite{lucas1981iterative}, and then estimating and verifying epipolar geometry using bundle adjustment~\cite{triggs1999bundle}.
Later works have scaled up the SfM pipeline using sequential algorithms, demonstrating results on hundreds or even thousands of images~\cite{snavely2006photo,furukawa2010towards,schoenberger2016mvs,schoenberger2016sfm,sarlin2019coarse}.

The advent of deep learning has augmented various stages of the classical SfM pipeline. Better feature descriptors~\cite{detone2018superpoint,simonyan2014learning,wang2020learning,yi2016lift,Dusmanu2019CVPR,Pautrat_2020_ECCV,revaud2019r2d2} and improved featured matching~\cite{sarlin2020superglue,choy2016universal,liu2010sift,truong2020glu,dusmanu2020multi}
have significantly outperformed their hand-crafted counterparts.
BA-Net~\cite{tang2018ba} and DeepSFM~\cite{wei2020deepsfm} have even replaced the bundle-adjustment process by optimizing over a cost volume. Most recently, Pixel-Perfect SfM~\cite{lindenberger2021pixel} uses a featuremetric error to post-process camera poses to achieve sub-pixel accuracy.

While these methods can achieve excellent localization, all these approaches are bottom-up: beginning with local features that are  matched across images. However, matching features requires sufficient overlap between images, which may not be possible given wide baseline views.
While our work also aims to localize camera poses given image sets, our approach fundamentally differs because it is top-down and does not rely on low-level correspondences.

\parhead{Simultaneous Localization and Mapping (SLAM).}
Related is the task of Monocular SLAM, which aims to localize and map the surroundings from a video stream. Indirect SLAM methods, similar to SfM, match local features across different images to localize the camera~\cite{rosinol2020kimera,ORBSLAM3_TRO,murORB2,mur2015orb}. Direct SLAM methods, on the other hand, define a geometric objective function to directly optimize over a photometric error~\cite{Zubizarreta2020,schops2019bad,davison2007monoslam,engel2016direct}.

There have also been various attempts to introduce deep learning into SLAM pipelines. As with SfM, learned feature descriptors and matching have helped improve accuracy on SLAM subproblems and increased robustness. End-to-end deep SLAM methods~\cite{zhou2018deeptam,newcombe2011dtam,wang2017deepvo,wang2020tartanvo} have improved the robustness of SLAM compared to classical methods, but have generally not closed the gap on performance.
One notable exception is the recent DROID-SLAM~\cite{teed2021droid}, which combines the robustness of learning-based SLAM with the accuracy of classical SLAM.

These approaches all assume \textit{sequential} streams and generally rely on matching or otherwise incorporating temporal locality between neighboring frames. We do not make any assumptions about the order of the image inputs nor the amount of overlap between nearby frames.

\parhead{Single-view Pose Prediction.}
The task of predicting a (6-DoF) pose from a single image has a long and storied history, the surface of which can barely be scratched in this section. Unlike relative pose between multiple images, the (absolute) pose given a single image is only well-defined if there exists a canonical coordinate system.
Most single-view pose prediction approaches therefore deal with a fixed set of categories, each of which has a canonical coordinate system defined \textit{a priori}~\cite{xiang2018posecnn,tekin18,Oberweger2018,chen2022occlusion,wang2019densefusion,Iwase2021repose,bukschat2020efficientpose,song2020hybridpose,novotny2019c3dpo,implicitpdf2021,kehl2017ssd,kendall2015posenet} or learned~\cite{sun2021canonical}.
Other methods that are category-agnostic take in a 3D mesh or point cloud as input, which provides a local coordinate system~\cite{wong2017segicp,Xiao2019PoseFromShape,zhang2020phosa,okorn2021zephyr}.

Perhaps most relevant to us are approaches that not only predict pose but also model inherent uncertainty in the pose prediction~\cite{brachmann2016uncertainty,kendall2016modelling,implicitpdf2021,okorn2020learning,corona2018pose,sundermeyer2018implicit,prokudin2018deep,gilitschenski2019deep,deng2019pose,deng2020self,mohlin2020probabilistic,manhardt2019explaining}. Like our approach, VpDR-Net~\cite{novotny2017learning} uses relative poses as supervision but still predicts absolute pose (with a unimodal Gaussian uncertainty model).
Implicit-PDF~\cite{implicitpdf2021} is the most similar approach to ours and served as an inspiration. Similar to our approach, Implicit-PDF uses a neural network to implicitly represent probability using an energy-based formulation which elegantly handles symmetries and multimodal distributions.
Unlike our approach, Implicit-PDF (and all other single-view pose prediction methods) predict \textit{absolute} pose, which does not exist in general for generic or novel categories. Instead, we model probability distributions over relative pose given pairs of images.

\parhead{Learning-based Relative Pose Prediction.} When considering generic scenes, prior works have investigated the task of relative pose prediction given two images. However, these supervised~\cite{Ummenhofer17demon} or self-supervised~\cite{Zhou_2017_CVPR,Yin_2018_CVPR,Mahjourian_2018_CVPR,vijayanarasimhan2017sfm,Lai21autoencoder} methods typically consider prediction of motion between consecutive frames and are not easily adapted to wide-baseline prediction. While some approaches have investigated wide baseline prediction~\cite{melekhov2017relative, balntas2018relocnet, rockwell20228}, regression-based inference can not effectively capture uncertainty unlike our energy-based model. Perhaps most similar to ours is DirectionNet~\cite{chen2021wide} which also predicts a camera distribution for wide baseline views. 
While DirectionNet only uses the expected value of the distribution and thus ignores symmetry, we take advantage of multimodal distributions  to improve our joint pose estimation.

\section{Method}

\begin{figure}[t]
    \centering
    \includegraphics[width=\textwidth]{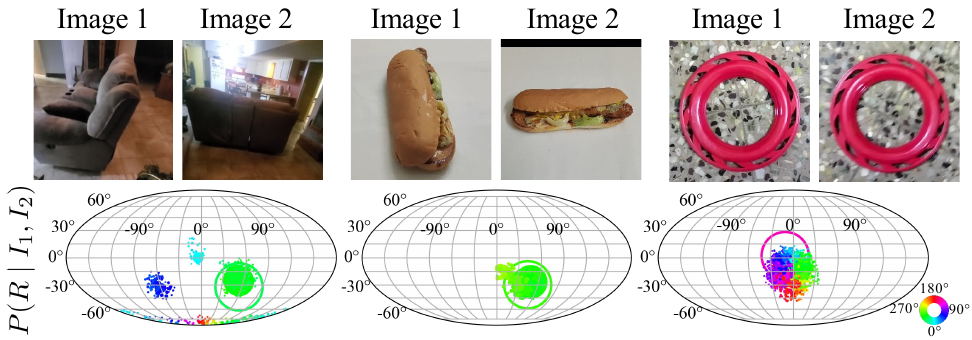}
    \caption{\textbf{Predicted conditional distribution of image pairs from unseen categories.} Here, we visualize the predicted conditional distribution of image pairs. Inspired by \cite{implicitpdf2021}, we visualize the rotation distribution~(\algoref{alg:pairwise}) by plotting yaw as latitude, pitch as longitude, and roll as the color. The size of each circle is proportional to the probability of that rotation. We omit rotations with negligible probability. The center of the open circle represents the ground truth. We can see that network predicts 4 modes for the couch images, corresponding roughly to 90 degree increments, with the greatest probability assigned to the correct 90 degree rotation. The relative pose of the hot dog is unambiguous and thus only has one mode. While the relative pose for the frisbee has close to no pitch or yaw, the roll remains ambiguous, hence the variety in colors. See the supplement for a visualization of how to interpret the relative rotations.
    }
    \label{fig:qual_pairwise}
\end{figure}

Given a set of $N$ images $\{I_1, \ldots I_N\}$ depicting a \emph{generic} object in-the-wild, we aim to recover a set of $N$ rotation matrices $\{R_1, \ldots, R_N\}$ such that rotation matrix $R_i$ corresponds to the viewpoint of the camera used to take image $i$.
Note that while we do not model translation, it can be easily initialized using object-facing viewpoints for 3D object reconstruction~\cite{lin2021barf,zhang2021ners} or a pose graph for SLAM~\cite{Carlone2015InitializationTF}.
We are primarily interested in settings with only sparse views and wide baselines. 
While bottom-up correspondence based techniques can reliably recover camera pose given dense views, they do not adapt well to sparse views with minimal overlap. We instead propose a prediction-based top-down approach that can learn and exploit the global structure directly.

The basic building block of our prediction system (visualized in \figref{fig:qual_pairwise}) is  a pairwise pose predictor that infers \textit{relative} camera orientations given pairs of images. However, symmetries in objects and possibly uninformative viewpoints make this an inherently uncertain prediction task.
To allow capturing this uncertainty, we propose an energy-based approach that models the \emph{multi-modal distribution} over relative poses given two images.


Given the predicted distributions over pairwise relative rotations, we show that these can be leveraged to induce a \emph{joint} distribution over the rotations. Starting with a greedy initialization, we present a coordinate-ascent approach that jointly reasons over and improves the set of inferred rotations. We describe our approach for modeling probability distributions over relative poses between two images in \secref{sec:method_pair}, and build on this in \secref{sec:method_joint} to recover a joint set of poses across multiple images. Finally, we discuss implementation details in \secref{sec:method_implementation}. 

\subsection{Estimating Pair-wise Relative Rotations}
\label{sec:method_pair}

\begin{wrapfigure}[10]{R}{0.6\textwidth}
\begin{minipage}{0.6\textwidth}
\vspace{-14mm}
\begin{algorithm}[H]
\footnotesize
\algnewcommand{\LineComment}[1]{\State \(\triangleright\) #1}
    \caption{\textbf{Pseudo-code for recovering a pairwise distribution}. We describe how to recover the distribution of the relative pose given images.}
    \label{alg:pairwise}
    \begin{algorithmic}
        \Procedure{PairwiseDistribution}{$I_1$, $I_2$}
            \State \texttt{queries} $\gets$\textsc{SampleRotationsUnif}(50000)
            \State \texttt{energies} $\gets f(I_1, I_2,$ \texttt{queries})
            \State \texttt{probs} $\gets$ \textsc{SoftMax}(\texttt{energies})
            \State \textbf{return} \texttt{queries}, \texttt{probs}
        \EndProcedure
    \end{algorithmic}
\end{algorithm}
\end{minipage}
\end{wrapfigure}

Given a pair of images depicting an arbitrary object, we aim to predict a distribution over the relative rotation corresponding to the camera transformation between the two views. As there may be ambiguities when inferring the relative pose given two images, we introduce a formulation that can model uncertainty.

\parhead{Energy-based Formulation.} We wish to model the conditional distribution over a relative rotation matrix $R$ given input images $I_1$ and $I_2$: $P(R \mid I_1, I_2)$.
Inspired by recent work on \textit{implicitly} representing the distribution over rotations using a neural network \cite{implicitpdf2021}, we propose using an energy-based relative pose estimator. More specifically, we train a network $f(R, I_1, I_2)$ that learns to predict the energy, or the unnormalized joint log-probability, $P(R, I_1, I_2) = \alpha \exp f(R, I_1, I_2) $  where $\alpha$ is the constant of integration. From the product rule, we can recover the conditional probability as a function of $f$:
\begin{equation}
\small{
    P(R \mid I_1, I_2) = \frac{P(R, I_1, I_2)}{P(I_1, I_2)} \approx \frac{\alpha \exp f(R, I_1, I_2)}{\sum_{R'} \alpha \exp f(R', I_1, I_2)} = \frac{ \exp f(R, I_1, I_2)}{\sum_{R'} \exp f(R', I_1, I_2)} }
    \label{eq:conditional}
\end{equation}
\noindent We marginalize over rotations to avoid having to compute $\alpha$ (see \algoref{alg:pairwise}), but note that the number of sampled rotations should be large for the approximation to be accurate. It is therefore important to use a lightweight network $f$ since it is queried once per sampled rotation in the denominator.

\parhead{Training.} We train our network by maximizing the log-likelihood of the conditional distribution, or equivalently minimizing the negative log-likelihood:
\begin{equation}
    \mathcal{L} = -\log P(R_1\T R_2 \mid I_1, I_2)
    \label{eq:loss_pairwise}
\end{equation}
where $R_1$ and $R_2$ are the ground truth poses of $I_1$ and $I_2$ respectively. Note that while the `absolute' poses $(R_1, R_2)$ are in an arbitrary coordinate system (depending on e.g.~SLAM system outputs), the relative pose $R_1\T R_2$ between two views is agnostic to this incidental canonical frame. Following \eref{eq:conditional}, we sample multiple candidate rotation matrices to compute the conditional probability. 

\parhead{Inference.} Recovering the optimal transformation from the pose of $I_1$ to $I_2$ amounts to optimizing $f$ over the space of rotations:
\begin{equation}
    R^* = \argmax_{R\in \mathbf{SO}(3)} P(R \mid I_1, I_2) = \argmax_{R\in \mathbf{SO}(3)} f(R, I_1, I_2)
\end{equation}
In practice, the loss landscape of $f$ is often un-smooth, so we find that sampling and scoring rotations based on $f$ to be more effective than gradient ascent.

We can also compute the conditional distribution of the relative rotation from $I_1$ to $I_2$ by sampling rotations over $\mathbf{SO}(3)$. The probability associated with each rotation can be computed using a softmax function, as described \algoref{alg:pairwise} and derived in \eref{eq:conditional}.
Inspired by \cite{implicitpdf2021}, we can visualize  the distribution of rotations by projecting the rotation matrices on a 2-sphere using pitch and yaw and coloring the rotation based on roll. See \figref{fig:qual_pairwise} and the supplement for sample results. 

\subsection{Recovering Joint Poses}
\label{sec:method_joint}

\begin{wrapfigure}[20]{R}{0.6\textwidth}
\vspace{-12mm}
\begin{minipage}{0.6\textwidth}
\begin{algorithm}[H]
\footnotesize
\algnewcommand{\LineComment}[1]{\State \(\triangleright\) #1}
    \caption{\textbf{Pseudo-code for joint inference using relative pose predictor}. We describe how to recover the joint poses given $n$ images via coordinate ascent.}
    \label{alg:method}
    \begin{algorithmic}
        \Procedure{CoordAsc}{Images $\{I_i\}_N$}
            \State $\{R_i\}_N \gets$ \textsc{InitializeRotations}($\{I_i\}_N$)
            \For{$t \in 1, \ldots, $ Num Iterations}
                \State $k \gets$ \textsc{RandomInteger}($N$)
                \LineComment{ $R_k'$ ($Q$$\times$3$\times$3): $Q$ replacements for $R_k$}
                \State $R_k' \gets$\textsc{SampleRotationsUnif}($Q$=250000)
                \State \texttt{energs} $\gets$ \textsc{Zeros}($Q$)
                \For{$i\in 1, \ldots, N$ \textbf{and} $i\not=k$}
                    \State $R \gets$  \textsc{Repeat}($R_i$, $Q$) \Comment{3$\times$3$\rightarrow$$Q$$\times$3$\times$3}
                    \State \texttt{energs} $\gets$ \texttt{energs} + $f(I_i, I_k, R^\top R_k')$
                    \State \texttt{energs} $\gets$ \texttt{energs} + $f(I_k, I_i, R_k'^\top R)$
                \EndFor
                \State $R_k \gets R_k'$[\textsc{ArgMax}(\texttt{energs})]
            \EndFor
        \EndProcedure
    \end{algorithmic}
\end{algorithm}
\end{minipage}
\end{wrapfigure}

In the previous section, we describe an energy-based relative pose predictor conditioned on pairs of images. Using this network, we recover a coherent set of rotations when given a set of images.

\parhead{Greedy Initialization.} Given predictions for relative rotations between every pair of images, we aim to associate each image with an absolute rotation. However, as the relative poses are invariant up to a global rotation, we can treat the pose of the first image as the identity matrix: $R_1 = I$. We note that the rotations for the other images can be uniquely induced given any $N-1$ relative rotations that span a tree.

\vspace{2mm}
\noindent \emph{Sequential Chain}. Perhaps the simplest way to construct such a tree is to treat the images as part of an ordered sequence. Given $R_1 = I$, all subsequent poses can be computed by using the best scoring relative pose from the previous image: $R_i = R_{i - 1} R^*_{(i-1)\shortrightarrow i}$, denoting $R_{i\shortrightarrow j}$ as the relative rotation matrix $R_i\T R_j$. However, this assumes that the images are captured sequentially (e.g. in a video) and may not be applicable for settings such as online marketplaces.

\noindent \emph{Maximum Spanning Tree}. We improve over the naive linear chain by recognizing that some pairs of images may produce more confident predictions. Given $N$ images, we construct a directed graph with $N\cdot (N-1)$ edges, where the weight of edge $(i, j) = P(R^*_{i\shortrightarrow j}|I_i, I_j)$. We then construct a Maximum Spanning Tree (MST) that covers all images with the most confident set of relative rotations.

\vspace{1mm}
\parhead{Reasoning over all images jointly.} Both of the previous methods, which select a subset of edges, do not perform any joint reasoning and discard all but the highest scoring mode for each pair of images. Instead, we can take advantage of our energy-based formulation to enforce global consistency.

Given our pairwise conditional probabilities, we can define a joint distribution over the set of rotations:  
\begin{equation}
    P\left(\{R_i\}_{i=1}^N \mid \{I_i\}_{i=1}^N\right) = \alpha \exp \Big( \sum_{(i, j) \in \mathcal{P}} f(R_{i\rightarrow j} \mid I_i, I_j) \Big)
    \label{eq:conditional_joint}
\end{equation}
where $\mathcal{P} = \{(i, j) \mid (i, j) \in [N]\times [N], i \not= j \}$ is the $N(N-1)$ set of pairwise permutations and $\alpha$ is the normalizing constant. Intuitively, this corresponds to the distribution modeled by a factor graph with a potential function corresponding to each pairwise edge.

\begin{figure}[t]
    \centering
    \includegraphics[width=\textwidth]{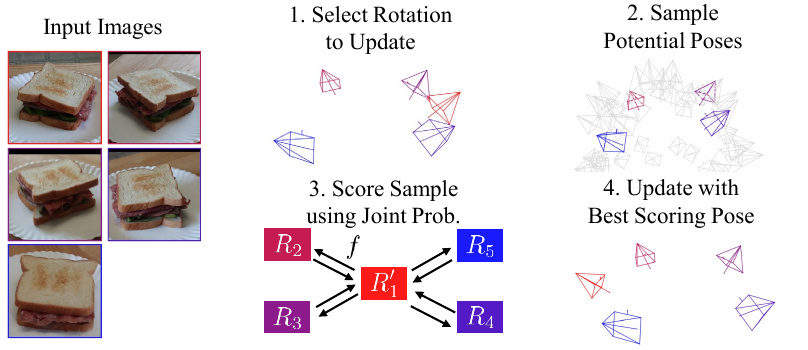}
    \caption{\textbf{Recovering Joint Poses with Coordinate Ascent.} Given a set of images $\{I_1, \ldots, I_N\}$, we initialize a set of corresponding poses $\{R_1, \ldots, R_N\}$. During each iteration of coordinate ascent, we: 1)~randomly select one pose $R_k$ to update (the red camera in this case); 2)~sample a large number (250k) of candidate poses; 3)~score each pose according to the joint distribution conditioned on the other poses and images ~\eref{eq:log_prob}; and 4)~update with the highest scoring pose. See \secref{sec:method_joint} for more detail.}
    \label{fig:coord_ascent}
\end{figure}

We then aim to find the most likely set of rotations $\{R_1, \ldots, R_N\}$ under this conditional joint distribution (assuming $R_1 = I$). While it is not feasible to analytically obtain the global maxima, we adopt an optimization-based approach and iteratively improve the current estimate. More specifically, we initialize the set of poses with the greedy MST solution and at each iteration, we randomly select a rotation $R_k$ to update. Assuming fixed values for $\{R_i\}_{i \neq k}$, we then search for the rotation $R_k$ under the conditional distribution that maximizes the overall likelihood. We show in supplementary that this in fact corresponds to computing the most likely hypothesis under the distribution $P(R_k' \mid \{R_i\}_{i \not = k}, \{I_i\}_i)$:
\begin{align}
    \log P(R_k' \mid \{R_i\}_{i \not = k}, \{I_i\}_i) = \sum_{i\not=k} \left( f(R_{i\rightarrow k'}, I_i, I_k) + f(R_{k'\rightarrow i}, I_k, I_i)\right) + C
    \label{eq:log_prob}
\end{align}
Analogous to our approach for finding the optimal solution for a single relative rotation, we sample multiple hypotheses for the rotation $R_k$, and select the hypothesis that maximizes \eref{eq:log_prob}. We find that this search-based block coordinate ascent helps us consistently improve over the initial solution while avoiding the local optima that a continuous optimization is susceptible to. We provide pseudo-code in \algoref{alg:method} and visualize one iteration of coordinate ascent in \figref{fig:coord_ascent}.

\subsection{Implementation Details}
\label{sec:method_implementation}

\parhead{Network Architecture.} We use a ResNet-50~\cite{he2016deep} with anti-aliasing~\cite{zhang2019shiftinvar} to extract image features.
We use a lightweight 3-layer MLP that takes in a concatenation of 2 sets of image features and a rotation matrix to predict energy. We use positional encoding~\cite{mildenhall2020nerf,tancik2020fourfeat} directly on flattened $3\times3$ rotation matrix, similar to~\cite{implicitpdf2021}. See the supplement for architecture diagrams.


\parhead{Number of Rotation Samples.} 
We use the equivolumetric sampling in~\cite{implicitpdf2021} to compute query rotations (37k total rotations) during training.
For each iteration of coordinate ascent, we randomly sample 250k rotation matrices. For visualizing distributions, we randomly sample 50k rotations.

\parhead{Runtime.} We train the pairwise estimator with a batch size of 64 images for approximately 2 days on 4 NVIDIA 2080TI GPUs. 
Inference for 20 images takes around 1-2 seconds to construct an MST and around 2 minutes for 200 iterations of coordinate ascent on a single 2080TI. Note that the runtime of the coordinate ascent scales linearly with the number of images.

\section{Evaluation}
\begin{figure}
    \centering
    \includegraphics[width=0.95\textwidth]{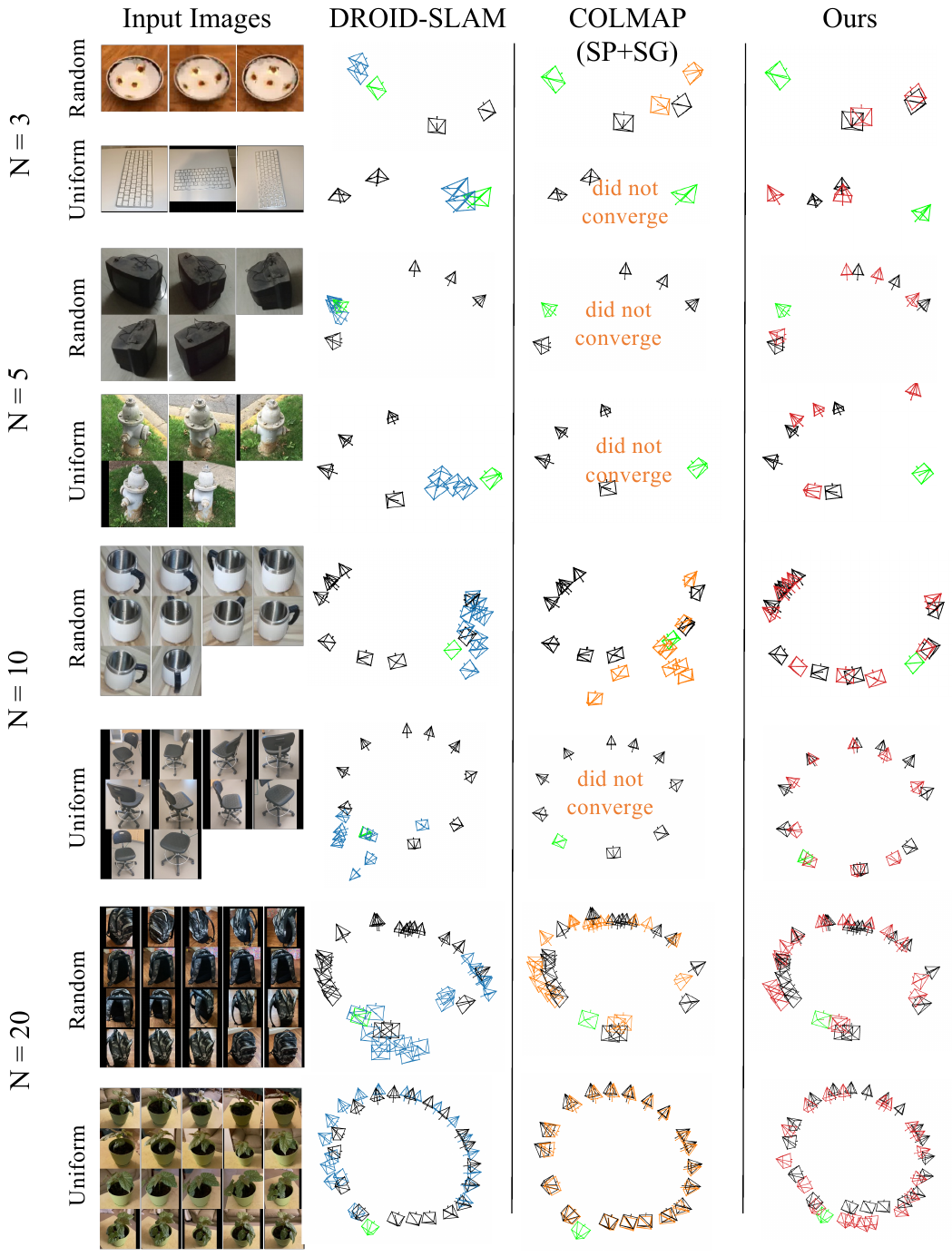}
    \caption{\textbf{Qualitative Comparison of Recovered Camera Poses with Baselines.} We visualize the camera poses (rotations) predicted by DROID-SLAM, COLMAP with SuperPoint/SuperGlue, and our method given sparse image frames. The black cameras correspond to the ground truth.
    We only visualize the rotations predicted by each method, and set the translation such that the object center is a fixed distance away along the camera axis.
    As the poses are agnostic to a global rotation, we align the predicted cameras across all methods to the ground truth coordinate system by setting the recovered camera pose for the first image to the corresponding ground truth (visualized in green).
    Odd rows correspond to randomly sampled image frames, while even rows correspond to uniformly-spaced image frames.
}
    \label{fig:joint_comparison}
\end{figure}

\begin{figure}[t]
    \centering
    \includegraphics[width=\textwidth]{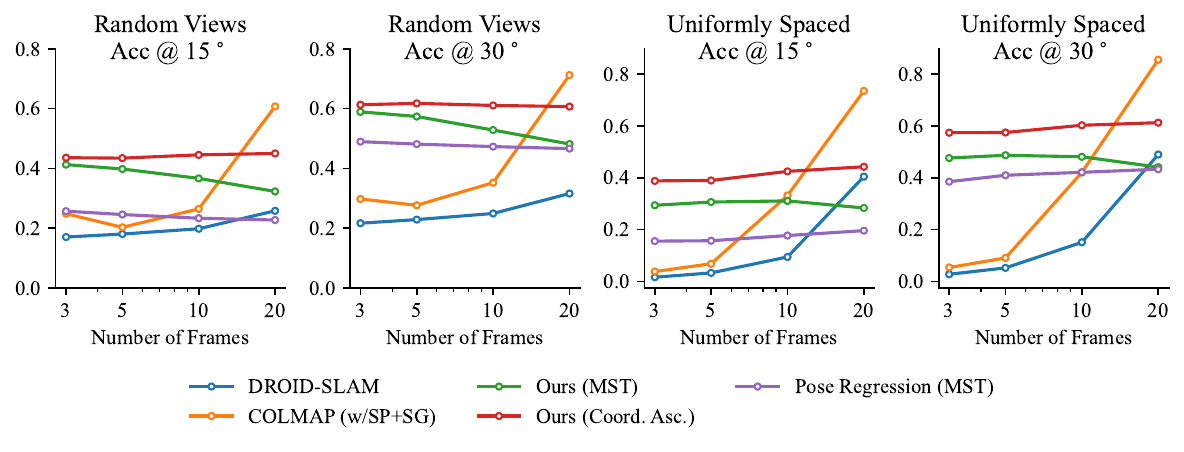}
    \caption{\textbf{Mean Accuracy on \textit{Seen} Categories.} We evaluate our approach against competitive SLAM (DROID-SLAM) and SfM (COLMAP with SuperPoint + SuperGlue) baselines in sparse-view settings. 
    We also train a direct relative rotation predictor (Pose Regression) that is not probabilistic and uses the MST generated by our method to recover joint pose.
    We consider both randomly sampling and uniformly spacing frames from a video sequence. We report the proportion of pairwise relative poses that are within 15 and 30 degrees of the ground truth, averaged over all seen categories. We find that our approach shines with fewer views because it does not rely on correspondences and thus can handle wide baseline views. Correspondence-based approach need about 20 images to begin to work.
    }
    \label{fig:seen_mean}
\end{figure}

\begin{figure}[t]
    \centering
    \includegraphics[width=\textwidth]{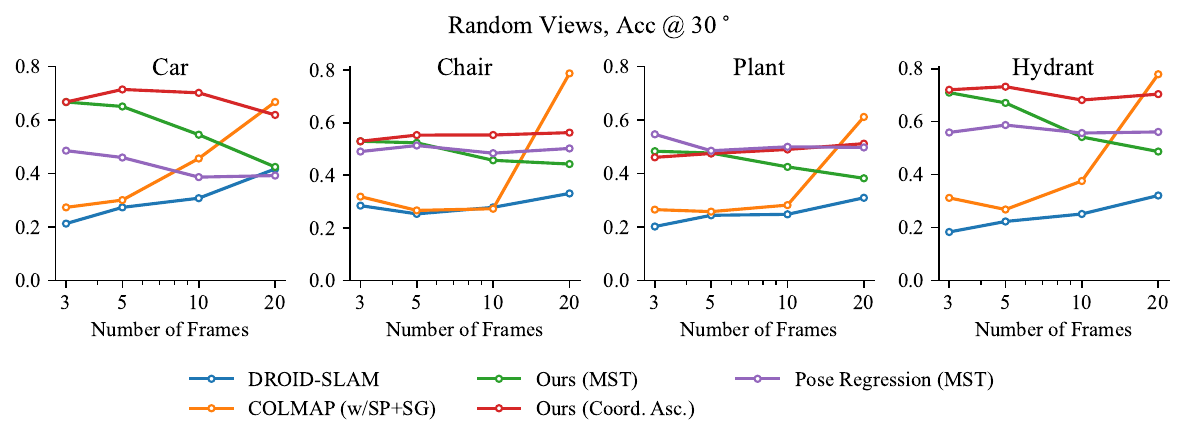}
    \caption{\textbf{Accuracy on Subset of \textit{Seen} Categories.} Here we compare all approaches on a representative subset of seen categories. We find that direct regression of relative poses (purple) struggles more on categories with symmetry (Car, Hydrant) than categories without symmetry (Chair, Plant), suggesting that multimodal prediction is important for resolving ambiguity.}
    \label{fig:seen_categories}
\end{figure}

\begin{figure}[t]
\centering
\begin{minipage}{.48\textwidth}
  \centering
  \includegraphics[width=\linewidth]{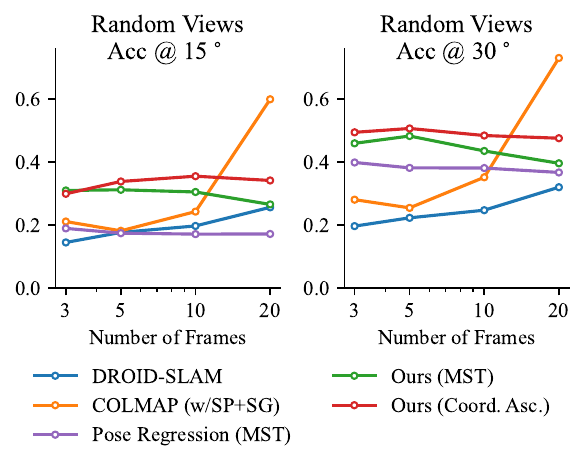}
  \captionof{figure}{\textbf{Mean Accuracy on \textit{Unseen} Categories.} We evaluate our approach on held out categories from CO3D.}
  \label{fig:unseen_mean}
\end{minipage}~~~%
\begin{minipage}{.48\textwidth}
  \centering
    \includegraphics[width=\textwidth]{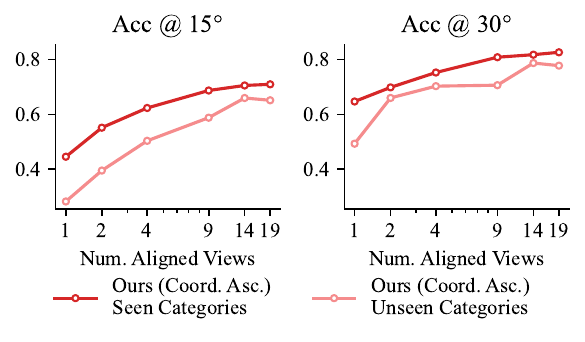}
     \captionof{figure}{\textbf{Novel View Registration.} Here, we evaluate the task of registering a new view given previously aligned cameras. We find that adding more views improves performance, suggesting that additional views reduce ambiguity.
    }
  \label{fig:hold_one_eval}
\end{minipage}
\end{figure}

\subsection{Experimental Setup}

\parhead{Dataset.}
We train and test on the Common Objects in 3D dataset (CO3D)~\cite{reizenstein21co3d}, a large-scale dataset consisting of turntable-style videos of 51 common object categories. 
We train on the subset of the dataset that has camera poses, which were acquired by running COLMAP~\cite{schoenberger2016sfm} over all frames of the video.

To train our network, we sample random frames and their associated camera poses from each video sequence. We train on 12,299 video sequences (from the \texttt{train-known} split) from 41 categories, holding out 10 categories to test generalization. We evaluate on 1,711 video sequences (from the \texttt{test-known} split) over all 41 trained categories (seen) as well as the 10 held out categories (unseen).
The 10 held out categories are: \texttt{ball}, \texttt{book}, \texttt{couch}, \texttt{frisbee}, \texttt{hotdog}, \texttt{kite}, \texttt{remote}, \texttt{sandwich}, \texttt{skateboard}, and \texttt{suitcase}. We selected these categories randomly after excluding some of the categories with the most training images.

\parhead{Task and Metrics.}
We consider the task of sparse-view camera pose estimation with $N =$ 3, 5, 10, and 20 images, subsampled from a video sequence. This is highly challenging, especially when $N\leq 10$, because the ground truth camera poses have wide baselines.

We consider two possible ways to select $N$ frames from a video sequence. First, we can randomly sample a set of $N$ indices per video sequence (Random). Alternatively, we can use $N$ uniformly-spaced frame indices (Uniform). We note that because CO3D video sequences are commonly taken in a turntable fashion, the uniformly spaced sampling strategy may be more representative of real world distributions of sparse view image sets. We report metrics on both task setups.

Because the global transformation of the camera poses is ambiguous, we evaluate each pair of relative rotations. For each of the $N(N-1)$ pairs, we compare the angular difference between the relative predicted rotation and the relative ground truth rotation using Rodrigues' formula~\cite{rodrigues1840lois}. We report the proportion of relative rotations that are within 15 and 30 degrees of the ground truth. We note that rotation errors within this range are relatively easy to handle by downstream 3D reconstruction tasks (See \figref{fig:ners} for an example).

\parhead{Baselines.} 
We compare against DROID-SLAM~\cite{teed2021droid}, a current state-of-the-art SLAM approach that incorporates learning in an optimization framework.
Note that DROID-SLAM requires trajectories and camera intrinsics.
Thus, we provide the DROID-SLAM baseline with sorted frame indices and intrinsics, but do not provide these to any other method.

We also compare with a state-of-the-art structure-from-motion pipeline that uses 
COLMAP~\cite{schoenberger2016sfm} with SuperPoint feature extraction~\cite{detone2018superpoint} and SuperGlue matching~\cite{sarlin2020superglue}.
We used the implementation provided by \cite{sarlin2019coarse}.
For instances for which COLMAP does not converge or is unable to localize some cameras, we treat the missing poses as identity rotation for evaluation. We note that DROID-SLAM also outputs approximate identity rotations when the optimization fails.

\parhead{Ablations.} 
In the spirit of learning-based solutions that directly regress pose, we train a network that predicts relative rotation directly given two images. Similar to our energy-based predictor, we pass the concatenated images features from a ResNet-50 into an MLP. We double the number of layers from 3 to 6 and add a skip connection to give this network increased capacity. Rotations are predicted using the 6D rotation representation~\cite{zhou2019continuity}. See the supplement for additional architecture details.
The relative pose regressor cannot directly predict poses for more than two images. To recover sets of poses from sets of images, we use the MST graph recovered by our method to link the pairs of relative rotations (we find that this performs better than linking the relative rotations sequentially).

To demonstrate the benefits of joint reasoning, we additionally report the performance of our method using the greedy Maximum Spanning Tree (MST) solution. The performance of the sequential solution is in the supplement.

\subsection{Quantitative Evaluation}
We evaluate all approaches on sparse-view camera pose estimation by averaging over all seen categories in \figref{fig:seen_mean}. We find that our approach outperforms all baselines for $N\leq 10$ images.
Correspondence-based approaches (DROID-SLAM and COLMAP) do not work until roughly 20 images, at which point image frames have sufficient overlap for local correspondences. However, real world multi-view data (e.g. marketplace images) typically have much fewer images. We find that coordinate ascent helps our approach scale with more image frames whereas the greedy maximum spanning tree accumulates errors with more frames. 

Directly predicting relative poses does not perform well, possibly because pose regression cannot model multiple modes, which is important for symmetrical objects. We visualize the performance for four categories in \figref{fig:seen_categories}. We find that the performance gap between our approach and direct regression is larger for objects with some symmetry (car, hydrant) than for objects without symmetry (chair, plant). Moreover, unlike our energy-based approach that models a joint distribution, a regression-based method does not allow similar joint reasoning.

We also test the generalization of our approach for \emph{unseen} categories in \figref{fig:unseen_mean}. We still find that our method significantly outperforms all other approaches with sparse view ($N\leq10$) even for never-before-seen object categories, indicating its ability to handle generic objects beyond training. The per-category evaluation for both seen and unseen categories are in the supplement.

\parhead{Novel View Registration.} In our standard SfM-inspired task setup, we aim to recover $N$ camera poses given $N$ images. Intuitively, adding images reduces ambiguity, but recovering additional cameras is also more challenging. To disambiguate between the two, we evaluate the task of registering new views given previously aligned images in \figref{fig:hold_one_eval}. Given $N + 1$ images, of which $N$ have aligned cameras, we use our energy-based regressor to recover the remaining camera (equivalent to one iteration of coordinate ascent). We find that adding images improves accuracy, suggesting that additional views can reduce ambiguity.

\subsection{Qualitative Results}
\begin{figure}[t]
    \centering
    \includegraphics[width=0.98\textwidth]{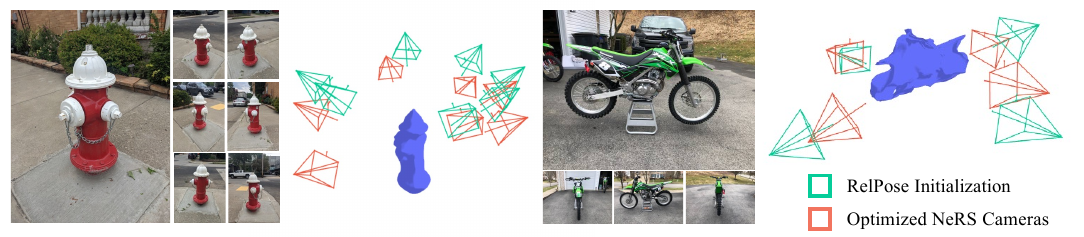}
    \caption{\textbf{Initializing 3D NeRS Reconstruction using Predicted Cameras.} NeRS~\cite{zhang2021ners} is a representative 3D reconstruction approach that takes noisy cameras as initialization and jointly optimizes object shape, appearance, and camera poses. We run our method with coordinate ascent on 7 input images of a fire hydrant and 4 input images of a motorbike
    to obtain the camera initialization (green), which we provide to NeRS. NeRS then finetunes the cameras (orange) and outputs a 3D reconstruction.
    }
    \label{fig:ners}
\end{figure}

We show qualitative results on the outputs of our pairwise predictor in \figref{fig:qual_pairwise}. The visualized distributions suggest that our model is learning useful information about symmetry and can model multiple modes even for unseen categories.

We visualize predicted camera poses for DROID-SLAM, COLMAP, and our method with coordinate ascent in \figref{fig:joint_comparison}.
Unable to bridge the domain gap from narrow baseline video frames, DROID-SLAM often gets stuck in the trajectory.
Although COLMAP sometimes fails to converge, it performs well for $N$=20. Our approach consistently outputs plausible interpretations but is unable to achieve \textit{precise} localization. See supplementary for visualizations on randomly selected sequences and more category-specific discussion.

We also validate that our camera pose estimations can be used for downstream 3D reconstruction. We use our camera poses to initialize NeRS~\cite{zhang2021ners}, a representative sparse-view surface-based approach that requires a (noisy) camera initialization. Using our cameras, we successfully reconstruct a 3D model of a fire hydrant from 7 images and a motorbike from 4 images in \figref{fig:ners}. 
Note that the camera pose initialization in the original NeRS paper was manually selected.

\section{Discussion}
We presented a prediction-based approach for estimating camera rotations given (a sparse set of) images of a generic object. Our energy-based formulation allows capturing the underlying uncertainty in relative poses, while also enabling joint reasoning over multiple images. We believe our system's robustness under sparse views can allow it to serve as a stepping stone for initializing (neural) reconstruction methods in the wild, but also note that there are several open challenges. First, our work reasoned about the joint distribution using only pairwise potentials and developing efficient higher-order energy models may further improve performance. Moreover, while we outperform existing techniques given sparse-views, the correspondence-driven methods are more accurate given a large number of views, and we hope future efforts can unify the two approaches. Finally, our approach may not be directly applicable to reasoning about camera transformations for arbitrary scenes as modeling camera translation would be more important compared to object-centric images.

\parhead{Acknowledgements.} We would like to thank Gengshan Yang, Jonathon Luiten, Brian Okorn, and Elliot Wu for helpful feedback and discussion. This work was supported in part by the NSF GFRP (Grant No. DGE1745016), Singapore DSTA, and CMU Argo AI Center for Autonomous Vehicle Research.

\clearpage
\section{Supplementary Materials}

\begin{figure}[h]
    \centering
    \includegraphics[width=\textwidth]{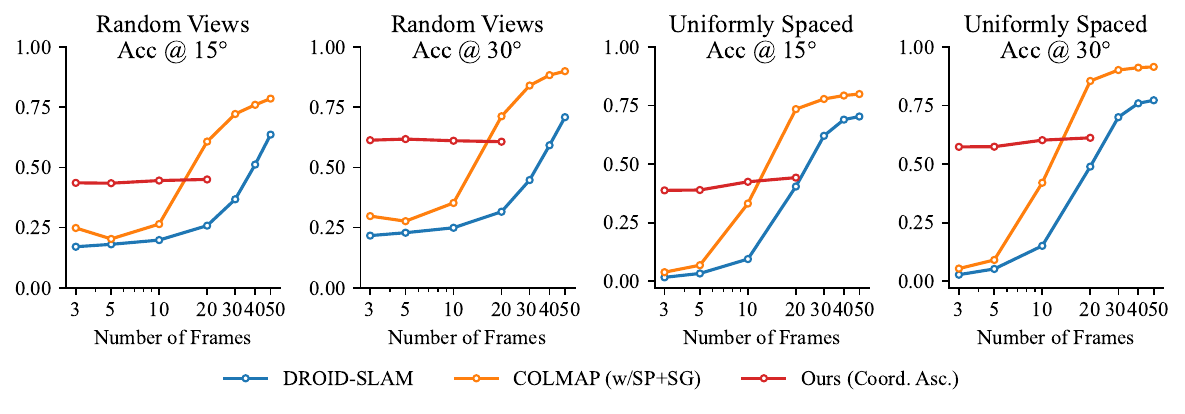}
    \caption{\textbf{Evaluation of correspondence-based approaches on large image sets (on ``Seen Categories" Split).} We evaluate the DROID-SLAM~\cite{teed2021droid} and COLMAP (with SuperPoint features and SuperGlue matching) baselines on much longer image sequences (N=30, 40, 50). We verify that these approaches, which rely on correspondences between images, can achieve good performance when the cameras baselines are narrow. Nonetheless, the poor performance at $N < 20$ suggests that there is a rich space for improving camera pose estimation in the low data regime, which is the setting that we target in our work.
    }
    \label{fig:supp_seen_all}
\end{figure}

\begin{figure}
    \centering
    \includegraphics[width=\textwidth]{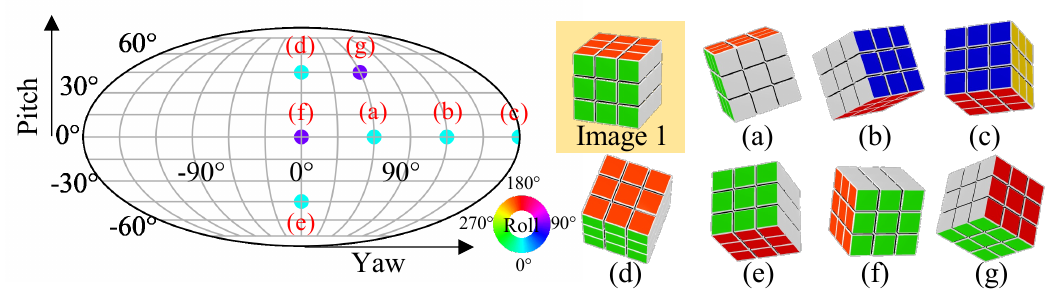}
    \caption{\textbf{Interpreting Relative Rotations using a 2-Sphere.} Given ``Image 1", we show how ``Image 2" would have appeared given different relative rotations. (a), (b), and (c) show relative rotations with 60\textdegree, 120\textdegree, and 180\textdegree~yaw respectively. (d) and (e) show relative rotations with 45\textdegree~and -45\textdegree~pitch respectively. (f) shows a relative rotation with just roll. (g) shows a relative rotation with all three components. We use a view-aligned coordinate system (See \figref{fig:view_aligned}) when computing relative rotations.
    Inspired by~\cite{implicitpdf2021}, we visualize the $\textbf{SO}(3)$ by projecting rotations onto a 2-sphere, with the x-axis representing yaw, y-axis representing pitch, and color representing roll.}
    \label{fig:interpreting_dist}
\end{figure}

\begin{figure}
    \centering
    \includegraphics[width=0.8\textwidth]{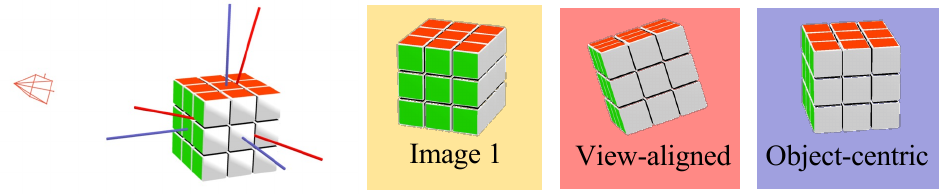}
    \caption{\textbf{View-aligned vs Object-centric Coordinate System.} 
    We compute relative rotations in a coordinate system (red axes on left) aligned with the camera (red wireframe on left). Relative rotations aligned to the camera  viewpoint can always be computed without reasoning about the object’s alignment with respect to the camera. While possibly more intuitive, relative rotations in the object coordinate system (blue axes on left) must be defined with respect to a canonical object pose and thus cannot be computed in general.
    On the right, we visualize a 60\textdegree~yaw relative rotation from Image 1 in the view-aligned coordinate system (red) and object-centric coordinate system (blue).}
    \label{fig:view_aligned}
\end{figure}

\begin{figure}
    \centering
    \includegraphics[width=\textwidth]{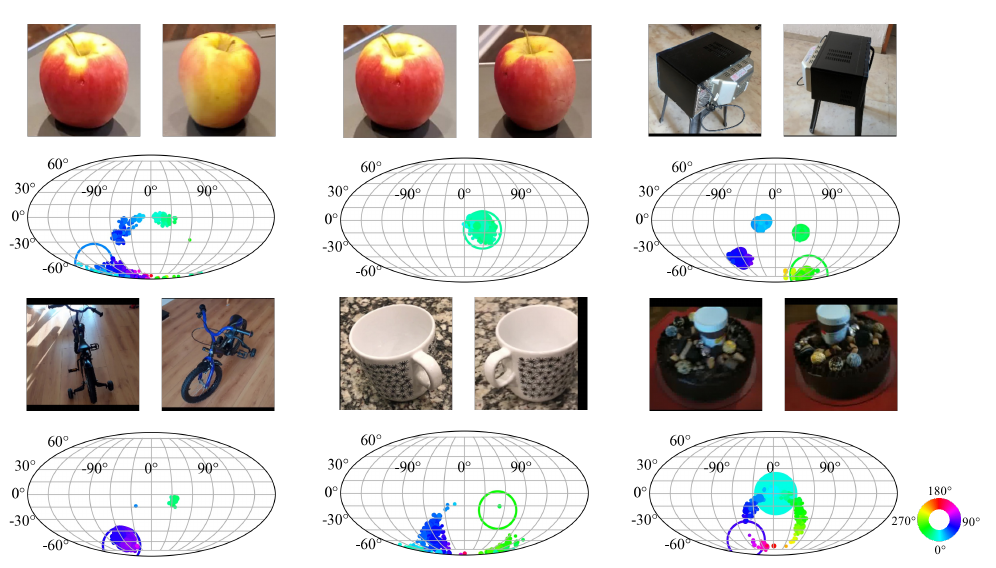}
    \caption{\textbf{Learned Pairwise Distributions on Seen Categories (Test Set).} Here we visualize the learned pairwise distributions for various pairs of images. \textit{Top~left:} The images correspond to opposite sides of the apple, so the relative pose is ambiguous. Our approach predicts a rotationally symmetric band of possible rotations. \textit{Top~middle:} The images have sufficient overlap such that the relative rotation is unambiguous and our method predicts a single mode for the apples. \textit{Top~right:} For rectangular objects such as microwaves, our approach often predicts 4 modes corresponding to each of the 90 degree rotations. \textit{Bottom~left:} Our approach predicts 2 modes for the bicycle because the first viewpoint is challenging. \textit{Bottom-middle:} Clashing foreground and background textures can be a challenge for our pairwise predictor. Even though the relative pose should be unambiguous, our method places low probability on the correct pose although it does recognize the rotational symmetry of the cup category. \textit{Bottom-right:} Unusual object appearances is another failure mode of our method, which defaults to placing high probability mass on the identity matrix. Our method does recognize the rotational symmetry of the cake category.}
    \label{fig:misc_pairwise}
\end{figure}

\begin{figure}
    \centering
    \includegraphics[width=0.9\textwidth]{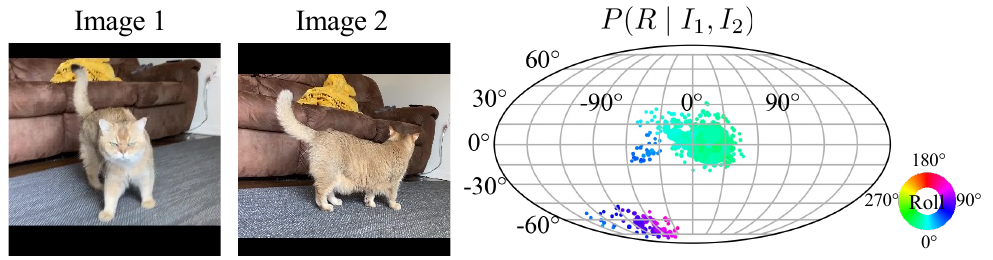}
    \caption{\textbf{Deformable Objects.} Existing SfM and SLAM pipelines often make assumptions about rigidity or appearance constancy in order for bundle adjustment to converge. Our method has no such requirements and can be run even on deformable objects. While the ground truth poses for these images of a cat are unknown, the relative rotation of the camera w.r.t the cat is roughly -90 degrees yaw with negative pitch while the relative rotation of the camera w.r.t. the couch has no pitch or yaw but some roll in the clockwise direction (green). Although our training data does not include dynamic or deformable objects, our network outputs plausible modes.}
    \label{fig:misc_objects}
\end{figure}

\begin{figure}
    \centering
    \includegraphics{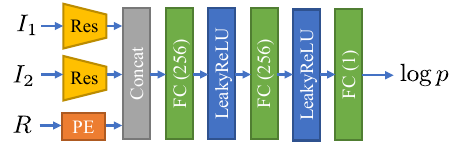}
    \caption{\textbf{Architecture Diagram for our Pairwise Energy Predictor.} We use a ResNet-50~\cite{he2016deep} with anti-aliasing~\cite{zhang2019shiftinvar} as our feature extractor. We directly apply positional encoding (8 bases)~\cite{tancik2020fourfeat} to the elements of the $3\times 3$ rotation matrix. We concatenate the image features and positionally encoded rotations into a feature vector ($2048 + 2048 + 2 \cdot 8 \cdot 9$), which we feed into an MLP that predicts energy (corresponding to unnormalized log probability).}
    \label{fig:arch_ours}
\end{figure}

\begin{figure}
    \centering
    \includegraphics[width=\textwidth]{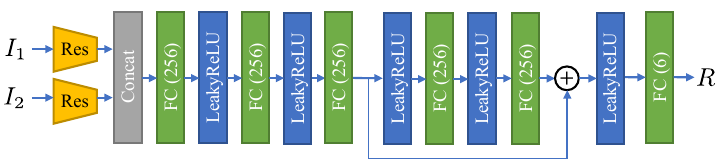}
    \caption{\textbf{Architecture Diagram for our Direct Pairwise Rotation Predictor.} For the direct rotation regression baseline, we still input the concatenated image features ($2048 + 2048$). To make the baseline more competitive, we increase the capacity of the MLP to have 6 layers and a skip connection. The network predicts the 6-D rotation representation~\cite{zhou2019continuity}.}
    \label{fig:arch_direct}
\end{figure}

\begin{figure}
    \centering
    \includegraphics[width=\textwidth]{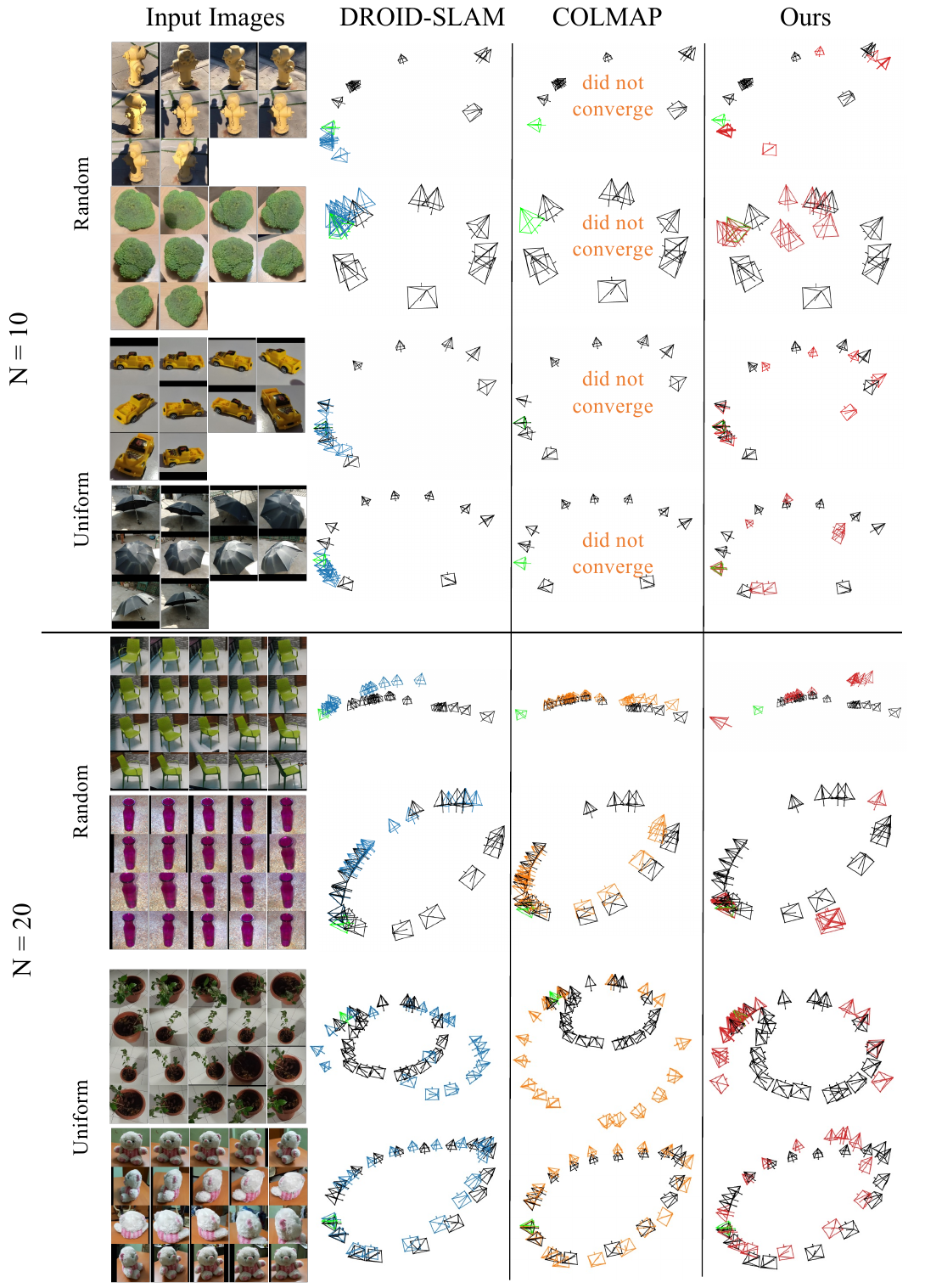}
    \caption{\textbf{Randomly selected Qualitative Results for Seen Categories.}}
    \label{fig:qual_random_seen}
\end{figure}

\begin{figure}
    \centering
    \includegraphics[width=\textwidth]{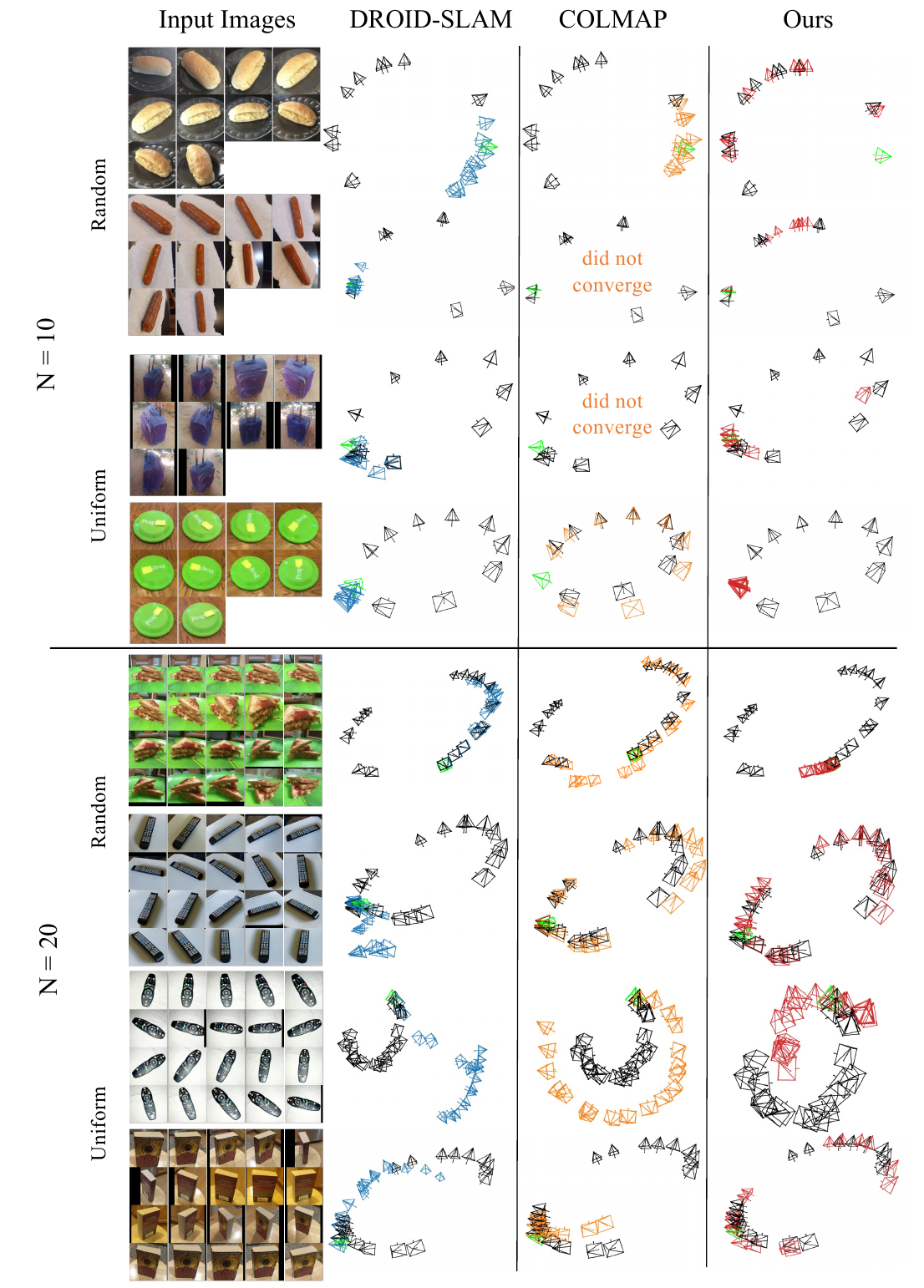}
    \caption{\textbf{Randomly selected Qualitative Results for Unseen Categories.}}
    \label{fig:qual_random_unseen}
\end{figure}

In this section, we show that maximizing the conditional distribution of an update to a hypothesis is equivalent to maximizing the joint likelihood in \secref{sec:conditional}. We evaluate ablations of our approach to validate the use of coordinate ascent vs gradient ascent and MST vs sequential loop in \tabref{tab:ablations}. To test the quality of our SLAM and SfM baselines, we also ran them with more image frames (narrower baseline) in \figref{fig:supp_seen_all}. We show per-category evaluations to compare performance across seen and unseen categories of CO3D in \tabref{tab:category_evaluation}. We provide a visualization of how to interpret the relative rotations in \figref{fig:interpreting_dist} and discuss the coordinate system in which we compute relative rotations in \figref{fig:view_aligned}. We discuss the learned symmetry modes as well as some failure modes in \figref{fig:misc_pairwise}. As a proof of concept, we use our energy-based predictor on a deformable object (cat) in \figref{fig:misc_objects}. 
We include architecture diagrams for our energy-based pairwise pose predictor in \figref{fig:arch_ours} and the direct pose predictor baseline in \figref{fig:arch_direct}.
Finally, we show qualitative comparisons between our approach and the correspondence-based baselines on \textit{randomly selected} sequences on both seen and unseen categories in \figref{fig:qual_random_seen} and \figref{fig:qual_random_unseen} respectively.

\setlength{\tabcolsep}{4pt}

\subsection{Derivation of Conditional Distribution for Coordinate Ascent}
\label{sec:conditional}

Given our pairwise conditional probabilities, the joint distribution over a set of rotations can be computed as:
\begin{equation}
\small{
    P\left(\{R_i\}_{i=1}^N \mid \{I_i\}_{i=1}^N\right) \propto P\left(\{R_i, I_i\}_{i=1}^N \right) = \alpha \exp \left( \sum_{(i, j) \in \mathcal{P}} f(R_{i\rightarrow j}, I_i, I_j) \right)
    }
    \label{eq:conditional_joint}
\end{equation}
\noindent where $\mathcal{P} = \{(i, j) \mid (i, j) \in [N]\times [N], i \not= j \}$. 

We are searching for the most likely set of rotations $\{R_1, \ldots, R_N\}$ under this joint distribution given images $\{I_1, \ldots, I_N\}$. 
For each iteration of coordinate ascent, we have our current most likely set of rotations $\{R_1, \ldots, R_N\}$ and wish to update $R_k$. If we fix all $\{R_i\}_{i\not=k}$, the only terms in $\mathcal{P}$ that can change are the ones involving $k$, and the rest can be folded into a scalar constant. Thus, searching for the rotation $R_k$ that maximizes the overall likelihood is equivalent to finding the most likely hypothesis under $P\left(R_k' \mid \{R_i\}_{i=1}^k, \{I_i\}_{i=1}^N\right)$:
\begin{align}
    \log P(R_k' \mid \{R_i\}_{i \not = k}, \{I_i\}_i) &= \sum_{(i, j) \in \mathcal{P}} f(R_{i\rightarrow j}, I_i, I_j) + C_1\\
    &=\sum_{i\not=k} \left( f(R_{i\rightarrow k'}, I_i, I_k) + f(R_{k'\rightarrow i}, I_k, I_i)\right) + C_2
\end{align}
\noindent This simplifies each iteration of coordinate ascent from a $\mathcal{O}(N^2)$ sum to a $\mathcal{O}(N)$ sum.

\begin{table}[h]
    \centering
    \begin{tabular}{lcccc}
        \toprule
        Acc @ $30^{\circ}$ & 3 & 5 & 10 & 20 \\  \midrule
        Ours (Sequential) & 0.50 & 0.48 & 0.42 & 0.39\\ %
        Ours (MST) & 0.52 & 0.50 & 0.47 & 0.43 \\
        Ours (Grad. Asc.) & 0.52 & 0.51 & 0.49 & 0.47\\ %
        Ours (Coord. Asc.) & \textbf{0.59} & \textbf{0.58} & \textbf{0.59} & \textbf{0.59} \\
        \bottomrule
        
    \end{tabular}
    \caption{\textbf{Ablations on Seen Categories in CO3D (Random Sequence Subsampling)}. 
    One way to convert a set of relative pose predictions to a coherent set of joint poses is by naively linking them together in a sequence (Sequential). We find that greedily linking them by constructing a maximum spanning tree (MST) performs slightly better since it incorporates that most confident relative rotation predictions. To make better use of our energy-based relative pose predictor, we tried directly running gradient ascent initialized from the MST solution and maximizing energy using ADAM (Grad. Asc.). 
    Because the loss landscape is non-smooth, we observe that it does not deviate much from the MST solution. We found the scoring-based block coordinate ascent (Coord. Asc.) to be the most effective.}
    \label{tab:ablations}
\end{table}

\begin{table}[t]
    \centering
    \footnotesize{
    \begin{tabular}{clrrrrr}
    \toprule
    & & \multicolumn{4}{c}{Acc. @ 30\textdegree~(\%)}\\
    & Category & 3 & 5 & 10 & 20 \\
    \midrule
\parbox[t]{2mm}{\multirow{28}{*}{\rotatebox[origin=c]{90}{Seen Categories}}}
& Apple &59&60&62&61 \\
& Backpack &63&58&59&57 \\
& Banana &67&54&63&55 \\
& Baseballbat &100&67&70&73 \\
& Baseballglove &48&56&56&55 \\
& Bench &69&75&68&66 \\
& Bicycle &62&61&63&62 \\
& Bottle &59&57&60&60 \\
& Bowl &80&75&77&80 \\
& Broccoli &55&54&51&51 \\
& Cake &46&47&47&54 \\
& Car &67&71&70&62 \\
& Carrot &60&64&63&65 \\
& Cellphone &69&78&72&69 \\
& Chair &53&55&55&56 \\
& Cup &55&56&54&51 \\
& Donut &52&44&51&51 \\
& Hairdryer &58&56&58&54 \\
& Handbag &66&63&62&61 \\
& Hydrant &72&73&68&70 \\
& Keyboard &72&73&74&74 \\
& Laptop &88&87&89&89 \\
& Microwave &56&65&55&58 \\
& Motorcycle &59&60&62&61 \\
& Mouse &68&70&69&67 \\
& Orange &52&52&51&49 \\
& Parkingmeter &22&27&23&22 \\
\bottomrule
    \end{tabular}\quad
    \begin{tabular}{clrrrrr}
    \toprule
    & & \multicolumn{4}{c}{Acc. @ 30\textdegree~(\%)}\\
    & Category & 3 & 5 & 10 & 20 \\
    \midrule
\parbox[t]{2mm}{\multirow{15}{*}{\rotatebox[origin=c]{90}{Seen Categories}}}
& Pizza &50&57&57&55 \\
& Plant &46&47&49&51 \\
& Stopsign &42&49&47&47 \\
& Teddybear &47&52&49&48 \\
& Toaster &76&75&71&73 \\
& Toilet &76&80&75&77 \\
& Toybus &63&70&72&71 \\
& Toyplane &43&57&48&51 \\
& Toytrain &81&73&75&75 \\
& Toytruck &71&69&68&68 \\
& Tv &78&83&87&86 \\
& Umbrella &58&60&54&55 \\
& Vase &58&55&55&51 \\
& Wineglass &51&46&46&47 \\
\cmidrule{2-6}
& Seen Mean &61&62&61&61 \\ 
\midrule
\parbox[t]{2mm}{\multirow{11}{*}{\rotatebox[origin=c]{90}{Unseen Categories}}}

& Ball &45&41&43&44 \\
& Book &51&49&49&47 \\
& Couch &42&58&39&35 \\
& Frisbee &55&49&40&38 \\
& Hotdog &58&61&50&49 \\
& Kite &28&23&27&24 \\
& Remote &64&58&65&66 \\
& Sandwich &37&41&41&42 \\
& Skateboard &56&64&64&65 \\
& Suitcase &59&61&67&63 \\
\cmidrule{2-6}
& Unseen Mean &49&51&48&48 \\
\bottomrule
    \end{tabular}
    }
    \caption{\textbf{Per-category Evaluation on CO3D with Random Sequence Sampling.} We find that rotationally symmetric objects (e.g. apple, orange, wineglass) tend to be challenging. We were surprised to find that bowls worked well, likely because the bowls in the CO3D dataset tend to have a lot of texture or even stickers. Objects with distinctive shapes (e.g. toilet, laptop) tend to be easier to orient. Note that some object categories have few instances for both training and testing (e.g. baseballbat, parkingmeter).}
    \label{tab:category_evaluation}
\end{table}

\clearpage

%
%

\bibliographystyle{splncs04}
\bibliography{egbib}

\end{document}